\pdfoutput=1

\documentclass[11pt]{article}

\usepackage[preprint]{acl}

\usepackage{times}
\usepackage{latexsym}

\usepackage{hyperref}
\usepackage{url}
\usepackage{tabularx}
\usepackage{booktabs}
\usepackage{multirow}
\usepackage{multicol}
\usepackage{soul}
\usepackage{graphicx}
\usepackage{xcolor,colortbl}
\definecolor{lightgray}{gray}{0.9}
\usepackage{bbding}
\usepackage{pifont}
\usepackage{diagbox}
\usepackage{longtable}
\usepackage{colortbl}
\usepackage{rotating}
\usepackage{wrapfig}
\usepackage[normalem]{ulem}

\usepackage[T1]{fontenc}

\usepackage[utf8]{inputenc}

\usepackage{microtype}

\usepackage{inconsolata}

\usepackage{graphicx}

\title{Pay Attention to Real World Perturbations! Natural Robustness Evaluation in Machine Reading Comprehension}

\author{Yulong Wu\textsuperscript{1}, Viktor Schlegel\textsuperscript{1, 2} and Riza Batista-Navarro\textsuperscript{1} \\
  \textsuperscript{1} Department of Computer Science, University of Manchester, United Kingdom \\
  \textsuperscript{2} Imperial College London, Imperial Global Singapore \\
  \texttt{\{yulong.wu, riza.batista\}@manchester.ac.uk} \\
  \texttt{v.schlegel@imperial.ac.uk}\\}

\begin{document}
\maketitle

\begin{abstract}

As neural language models achieve human-comparable performance on Machine Reading Comprehension (MRC) and see widespread adoption, ensuring their robustness in real-world scenarios has become increasingly important. Current robustness evaluation research, though, primarily develops synthetic perturbation methods, leaving unclear how well they reflect real life scenarios. Considering this, we present a framework to automatically examine MRC models on naturally occurring textual perturbations, by replacing paragraph in MRC benchmarks with their counterparts based on available Wikipedia edit history. Such perturbation type is \textit{natural} as its design does not stem from an arteficial generative process, inherently distinct from the previously investigated synthetic approaches. In a large-scale study encompassing \textsc{SQuAD} datasets and various model architectures we observe that natural perturbations result in performance degradation in pre-trained encoder language models. More worryingly, these state-of-the-art Flan-T5 and Large Language Models (LLMs) inherit these errors, with the largest observed drop reaching 28.28\%. Further experiments demonstrate that our findings generalise to natural perturbations found in other more challenging MRC benchmarks such as \textsc{DROP} and \textsc{HotpotQA}. In an effort to mitigate these errors, we show that robustness to natural perturbations can be improved through adversarial training for encoder-only models or through in-context demonstrations of perturbed instances for LLMs, although a more generalisable and effective defence strategy remains to be developed.
\end{abstract}

\section{Introduction}
\label{sec:Introduction}

Transformer-based pre-trained language models demonstrate remarkable efficacy in addressing questions based on a given passage of text, a task commonly referred to as Machine Reading Comprehension (MRC) \citep{devlin-etal-2019-bert, NEURIPS2020_1457c0d6, he2021deberta, wei2022finetuned, touvron2023llama, openai2024gpt4}. Despite these advancements, high-performing MRC systems are known to succeed by exploiting shortcuts in existing benchmark datasets rather than truly demonstrating understanding of the passage and question, thereby exhibiting a lack of robustness to various types of test-time perturbations. Such phenomenon is observed not only in encoder-only and encoder-decoder models~\citep{ho2023survey, Schlegel_Nenadic_Batista-Navarro_2023}, but also in state-of-the-art (SOTA) Large Language Models (LLMs)~\citep{levy-etal-2023-guiding, FANG2023200287, gupta-etal-2024-whispers}.

\begin{figure}[t!]
    \centering
    \includegraphics[width=\columnwidth]{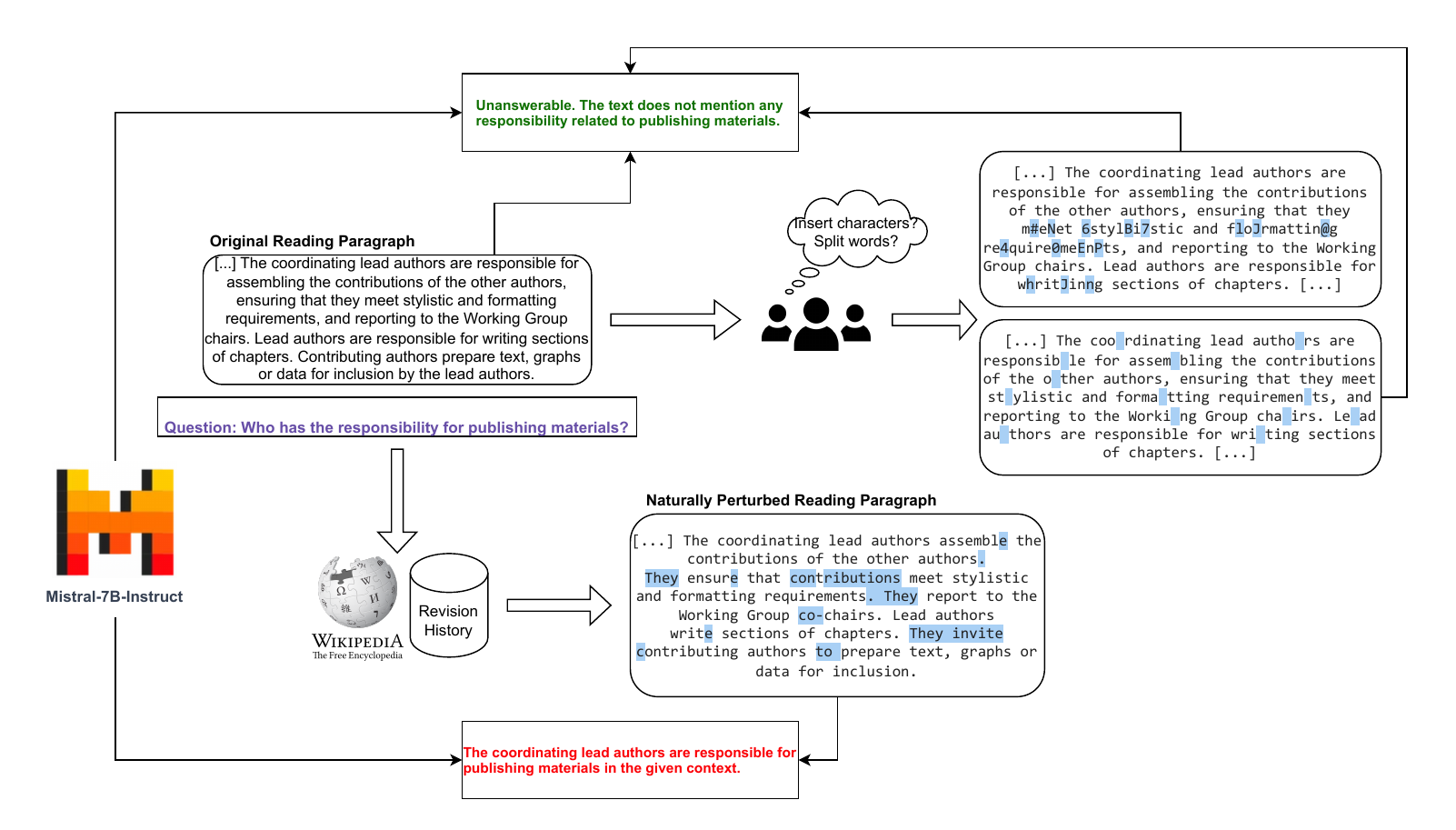}
    \caption{Given a reading paragraph, we extract and use Wikipedia revision history to construct its naturally perturbed version for a more realistic robustness evaluation (Bottom), rather than relying on a set of synthetic methods (Top). While~\texttt{Mistral-7B-Instruct-v0.2} generates the correct answer for both the original and synthetically perturbed passages, it fails under natural perturbation.}
    \label{fig:demonstration}
\end{figure}



Extensive textual perturbation methods have been developed to reveal the vulnerabilities of MRC models to various linguistic challenges~\citep{ribeiro-etal-2018-semantically, jiang-bansal-2019-avoiding, welbl-etal-2020-undersensitivity, tan-etal-2020-morphin, tan-joty-2021-code, Schlegel_Nenadic_Batista-Navarro_2021, cao-etal-2022-tasa, tran-etal-2023-impacts}. Despite the insights they provide, all of them craft perturbations in a synthetic manner—that is, based on hypothesised manipulation strategies (Figure~\ref{fig:demonstration} Top)—which may not accurately reflect the types of challenges encountered in real-world scenarios. As a result, there is a risk of overlooking the genuine weaknesses of reading comprehension systems when deployed in practical settings, potentially hindering efforts to improve their reliability in real applications.

To counteract this issue, in this paper, we develop a framework to inject textual changes that arise in real-world conditions into MRC datasets and audit how well contemporary language models perform under such perturbations. We deem them as \textit{natural} because the perturbation process does not involve any artificial manipulation, in line with the definitions by~\citet{belinkov2018synthetic, Hendrycks_2021_CVPR, Pedraza2022-yo, Agarwal_2022_CVPR, le-etal-2022-perturbations} (Figure \ref{fig:demonstration} Bottom).
Results of robustness evaluation are therefore more representative of real-world applications. Similar to~\citet{belinkov2018synthetic}, our approach utilises Wikipedia revision histories as the source of natural perturbations, given that the differences between revisions authentically capture the textual modifications made by human editors in the real world.
Despite this, significant differences exist in the perturbation construction methodology between us. Perturbation in~\citep{belinkov2018synthetic} is restricted to single word replacements and applied on non-English source-side sentences in machine translation. In detail, they build a look-up table of possible lexical replacements by harvesting naturally occurring errors (typos, misspellings, etc.) from available corpora of French/German Wikipedia edits~\citep{max-wisniewski-2010-mining, zesch-2012-measuring}. Afterwards, they replace every word in the source-side sentences with an error if one exists in the look-up table. Different from \citep{belinkov2018synthetic}, our approach does not restrict the perturbation level and utilise English Wikipedia. By comparing the variances between each adjacent revision, we identify perturbed versions for each Wikipedia reading passage in the original MRC benchmarks (if it exists). This enables us to capture more comprehensive and critical natural perturbation patterns (see Section \ref{subsec:Error Analysis}) that can not be possible to capture in \citep{belinkov2018synthetic}.
Our perturbation method only alter the reading context, while the questions and ground truth answers remain unchanged.


With the established framework, we conduct extensive experiments on six datasets, evaluating forty-two models, including recently proposed LLMs. Experimental results on Stanford Question Answering Dataset (\textsc{SQuAD}) \citep{rajpurkar-etal-2016-squad, rajpurkar-etal-2018-know} indicate that natural perturbations encompass rich linguistic variations and can lead to failures in the encoder-only models, while humans are almost undeterred by their presence. Crucially, these errors also transfer to larger and more powerful models, such as \texttt{Flan-T5} and SOTA LLMs, with performance drops ranging from $4.4$\% to $28.28$\%. These findings also generalise to other and more challenging MRC benchmarks (e.g., \texttt{Mistral-7B-Instruct-v0.2}'s $5.25$\% decrease on~\textsc{BoolQ}~\citep{clark-etal-2019-boolq} and \texttt{Llama-3.2-1B-Instruct}'s $9.98$\% decline on~\textsc{DROP}~\citep{dua-etal-2019-drop}), emphasising the harmful effects of natural perturbations. Adversarial re-training/in-context demonstration with either naturally or synthetically perturbed MRC instances can enhance the robustness against natural perturbations, with the latter sometimes providing greater benefits. However, there is still ample room for improvement, calling for better defense strategies.

The contributions of this paper are as follows:

\begin{itemize}
    \item A framework—based on Wikipedia revision history—for studying model robustness under real-world natural perturbations. This is relevant, even in the LLM era, as our framework can be applied to any other tasks with input from Wikipedia and also to any types of models.
    \item Perturbed datasets for six diverse MRC tasks. Two \textsc{SQUAD} challenge sets derived from error analysis of encoder-only models, on which SOTA LLMs struggle, even without being involved in the creation in any capacity.
    \item Empirical demonstration of the validity of natural perturbations across both encoder-only models and LLMs, their characterisation by different linguistic phenomena and their harmful effects on diverse model architectures across benchmarks generated with the proposed framework.
    \item Showcasing adversarial re-training with natural or, especially, synthetic perturbations, as well as adversarial in-context demonstrations as a way to enhance the robustness of encoder-only models and LLMs, respectively, against natural perturbations.
\end{itemize}

\section{Related Work}
\label{sec:Related Work}

\begin{figure*}[t!]
    \centering
    \includegraphics[width=\textwidth]{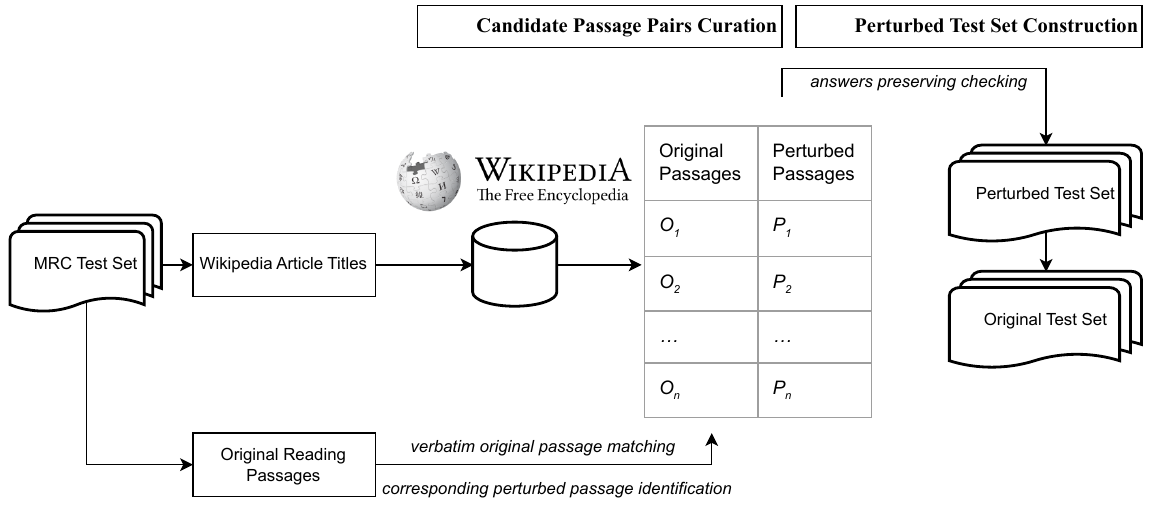}
    \caption{Process of generating naturally perturbed MRC test sets.}
    \label{fig:rm-updated}
\end{figure*}

\paragraph{Robustness Evaluation in MRC} A typical approach to evaluate the robustness of MRC models is via test-time perturbation. This line of research develops different perturbation methods as attacks, such as
adversarial distracting sentence addition \citep{jia-liang-2017-adversarial, tran-etal-2023-impacts}, low-level attacks~\citep{eger-benz-2020-hero}, word substitution \citep{wu-etal-2021-evaluating}, character swap \citep{si-etal-2021-benchmarking}, entity renaming \citep{yan-etal-2022-robustness} and paraphrasing \citep{gan-ng-2019-improving, lai-etal-2021-machine, wu-etal-2023-machine}. Our work also fits within the category of test-time perturbation, but differs from previous works in that we introduce perturbations that naturally occur in real-world scenarios, therefore contributing to a more practical robustness test. 


\paragraph{Natural Perturbation for Robustness Assessment} Compared with deliberately crafting the perturbed instances, the study of natural perturbation is quite under-explored. In the computer vision domain, researchers find that real-world clean images without intentional modifications can confuse deep learning models as well, terming them as natural adversarial examples \citep{Hendrycks_2021_CVPR, Pedraza2022-yo}. Similarly, in the field of Natural Language Processing (NLP), naturally occurring perturbations extracted from human-written texts can also degrade model performance in tasks such as machine translation \citep{belinkov2018synthetic} and toxic comments detection \citep{le-etal-2022-perturbations}.
Motivated by these, we attempt to harvest natural perturbations from available Wikipedia revision histories and utilise them to modify the original MRC instances.
\uline{To the best of our knowledge, we are the first to investigate MRC model robustness under real natural perturbations.}




\section{Natural Perturbation Pipeline}
\label{sec:Natural Perturbation Pipeline}





We design a pipeline to automatically construct label-preserving stress MRC test sets with noises that occur in real-world settings by leveraging Wikipedia revision histories (Figure \ref{fig:rm-updated}). Our approach comprises two modules: \textit{candidate passage pairs curation} and \textit{perturbed test set construction}.


\paragraph{Candidate passage pairs curation.} For each English Wikipedia article within the development set\footnote{Since not all test sets are public, we apply natural perturbations to the development sets. For simplicity, we use the term ``test set'' throughout.} of
MRC datasets, we systematically extract its entire revision histories and preprocess them, including the removal of markups and the segmentation of content. Subsequently, we obtain the content differences between each current revision and the previous adjacent one, identifying three distinct editing patterns: addition, deletion, and modification.
In the case of an edit falling within the modification pattern, we retain the paragraph from the prior version as the \textit{original} and the corresponding one from the current version as the \textit{perturbed}, provided both paragraphs exceed $500$ characters\footnote{This threshold setting adheres to the methodology employed in the collection of SQuAD 1.1 \citep{rajpurkar-etal-2016-squad}.}.

\paragraph{Perturbed test set construction.} To generate the naturally perturbed test set, we begin by acquiring all reading passages from the development set of each MRC dataset and identifying their entries in the collection of previously extracted candidate original passages, along with the corresponding perturbed counterparts. Subsequently, for the matched original passages with a single occurrence, we keep them and the corresponding perturbed passages; whereas for those with multiple occurrences, we randomly select one instance for each and extract its perturbed version. After obtaining the perturbed reading passages, we retain only those with at least one question where all annotated ground truth answers (or all plausible answers for the unanswerable question) can still be located within the perturbed context, resulting in the \emph{Perturbed} test set. For the sake of comparison, we also construct an \emph{Original} version of the test set keeping only the original passages and questions corresponding to those that were included in the \emph{Perturbed} version.

\section{Experiment Setup}
\label{sec:Experiments Setup}

\subsection{Datasets}
\label{subsec:Datasets}




We use six English MRC datasets: \textsc{SQuAD 1.1}~\citep{rajpurkar-etal-2016-squad}, \textsc{SQuAD 2.0}~\citep{rajpurkar-etal-2018-know}, \textsc{BoolQ}~\citep{clark-etal-2019-boolq}, \textsc{DROP}~\citep{dua-etal-2019-drop}, \textsc{HotpotQA} (distractor)~\citep{yang-etal-2018-hotpotqa} and \textsc{TyDi QA} (gold passage task in English)~\citep{clark-etal-2020-tydi}.
These are chosen
as their reading passages are sourced from Wikipedia, thereby enabling the utilisation of Wikipedia editing histories to generate the naturally perturbed test set.

\subsection{Models}
\label{subsec:Models}

Our evaluation study involves
MRC models across three different types: encoder-only, encoder-decoder, and decoder-only. Under the encoder-decoder and decoder-only model evaluation settings, we reframe
MRC as the text generation task based on the given context and question. Access to and experimentation with all models are possible via the use of the HuggingFace's \textit{Transformers} library \citep{wolf-etal-2020-transformers}, the vLLM library~\citep{10.1145/3600006.3613165}, two 80GB Nvidia A100 GPUs and the OpenAI ChatGPT API.

\paragraph{Encoder-only:} We select \texttt{BERT} \citep{devlin-etal-2019-bert} and its various variants for evaluation, including \texttt{DistilBERT} \citep{Sanh2019}, \texttt{SpanBERT} \citep{joshi-etal-2020-spanbert}, \texttt{RoBERTa} \citep{DBLP:journals/corr/abs-1907-11692}, \texttt{ALBERT} \citep{Lan2020ALBERT:} and \texttt{DeBERTa} \citep{he2021deberta}. Some of these model types also come with different variations, such as size (e.g., \emph{base} and \emph{large} for \texttt{RoBERTa}), versions (e.g., \emph{v1} and \emph{v2} for \texttt{ALBERT}) and whether the input text is cased or not (e.g., \emph{cased} and \emph{uncased} for \texttt{BERT}), all of which are included in the evaluation. We fine-tune these encoder-only pre-trained language models on the training set of the two \textsc{SQuAD} datasets \citep{rajpurkar-etal-2016-squad, rajpurkar-etal-2018-know} and evaluate them on the constructed original and perturbed test sets. Model details and the hyperparameters used in model fine-tuning are shown in Appendix \ref{sec:Encoder-only Model Parameters and Hyperparameters for Fine-tuning}.

\paragraph{Encoder–Decoder:} Instruction finetuning has been demonstrated to be effective in enhancing zero-shot performance of pretrained language models, resulting in the development of Finetuned Language Net (FLAN) \citep{wei2022finetuned}. In this work, we use the instruction-finetuned version of T5 model class, specifically the \texttt{Flan-T5} \citep{chung2022scaling}, available in sizes ranging from \emph{small} ($80$M), \emph{base} ($250$M), \emph{large} ($780$M) to \emph{xl} ($3$B). During evaluation, we utilise the instruction templates from MRC task collection in open-sourced FLAN repository and report the model performance as the average of those obtained across the employed templates. Refer to Appendix \ref{sec:Instruction Templates for Flan-T5 Evaluation} for various instruction templates used for the evaluation on the test sets with the format as the two \textsc{SQuAD} datasets.

\paragraph{Decoder-only:} There is an exponential increase of pre-trained
generative LLMs and their fine-tuned chat versions, inspired by the remarkable success of ChatGPT \citep{bang-etal-2023-multitask}.
Therefore, our experiments incorporate a broad range of recently proposed language model families, including \texttt{GPT 3.5 Turbo}, \texttt{GPT-4o}~\citep{openai2024gpt4ocard}, \texttt{Gemma}~\citep{Mesnard2024GemmaOM}, \texttt{Gemma 2}~\citep{Riviere2024Gemma2I},
\texttt{Llama 2}~\citep{touvron2023llama}, \texttt{Llama 3} and \texttt{Llama 3.1}~\citep{dubey2024llama3herdmodels}, \texttt{Llama 3.2},
\texttt{Mistral}~\citep{jiang2023mistral}, \texttt{OLMo}~\citep{groeneveld-etal-2024-olmo}, \texttt{Qwen2.5}~\citep{qwen2025qwen25technicalreport}, \texttt{Falcon}
\citep{almazrouei2023falcon}, \texttt{Falcon3}~\citep{Falcon3}, and \texttt{DeepSeek LLM}
\citep{deepseekai2024deepseekllmscalingopensource}. The zero-shot prompts designed for soliciting their responses are presented in Appendix~\ref{sec:MRC Prompts}.

\subsection{Evaluation Metrics}
\label{subsec:Evaluation Metrics}

In line with existing literature, we choose the
(instance-averaged) Token-F1 score to assess the performance of both encoder-only and encoder-decoder models~\citep{rajpurkar-etal-2016-squad}, as on \textsc{SQuAD}-style test sets, they are optimised to output the shortest continuous span from the context as the answer (or predict the question as unanswerable) during inference.
However, 
the outputs of the decoder-only models do not consistently adhere to the instruction due to their conversational style, rendering
F1 unsuitable for evaluation. Consequently,
we employ a more lenient metric, namely Inclusion Match (IM), which measures whether the response of the model contains any of the ground truth answers~\citep{bhuiya-etal-2024-seemingly}. Furthermore, if the model's output includes phrases such as ``I cannot answer this/the question'' or ``unanswerable''\footnote{We collate a collection of such phrases by manually examining the decoder-only models' outputs (Check Appendix \ref{sec:Indicators of Unanswerable} for the full set).}, we deem that the model believes the question is not answerable.
Model robustness is quantified by measuring the relative variation in performance (as reflected in the F1 or IM) under natural perturbations.


\section{MRC under Natural Perturbation}
\label{sec:MRC under Natural Perturbation}

In this section, we present and discuss the results of our experiments. We first evaluate encoder-only models on \textsc{SQuAD} to establish a baseline evaluation of model behaviour under natural perturbations. While neither represents the current SOTA, \textsc{SQUAD}'s simplicity, the stable, super-human performance of encoder-only models, and it's advantage of being free from benchmark leakage impact enable a focused and controlled examination of perturbation effects (Section \ref{subsec:Are Encoder-only MRC Models Resilient to Natural Perturbation?}), error sources (Section \ref{subsec:Error Analysis}), and adversarial instance validity (Section \ref{subsec:Validity of Nature Adversarial Examples}). Then, we investigate the transferability of errors from encoder-only models to other architectures, showing both \textsc{Flan-T5} and LLMs carry these errors significantly (Section~\ref{subsec:Can Errors from Encoder-only Models Affect Other Architectures?})~\footnote{To supplement, we further evaluate the full test set on \textsc{Flan-T5} and several LLMs, and measure the transferability of adversarial examples across all model architectures (Appendix \ref{sec:Impact of the Complete Set of Perturbed Instances on Encoder-Decoder and Decoder-Only Architectures}).}. We finally generalise the findings from the baseline evaluation to SOTA LLMs and other more complex datasets (Section~\ref{subsec:Do Our Findings Generalise to Other MRC Datasets?}).

\subsection{Are Encoder-only Models Resilient to Natural Perturbation?}
\label{subsec:Are Encoder-only MRC Models Resilient to Natural Perturbation?}

Table~\ref{tab:natural_encoder} presents the relative F1 change for encoder-only MRC models on the naturally perturbed~\textsc{SQuAD} test set. It can be seen that overall, the performance of all the examined models decreases, indicating that \textbf{\textit{encoder-only MRC models suffer from natural perturbation}}. Nonetheless, the performance decline of all models is not at a significant level, with the biggest drop being only $3.06$\%. This suggests that these models also exhibit considerable robustness to natural perturbation.

\begin{table}[htb!]
\centering
\resizebox{\columnwidth}{!}{
\begin{tabular}{lccc}
    \toprule
    \textbf{Victim} & \textbf{\textsc{SQuAD 1.1}} & \multicolumn{2}{c}{\textbf{\textsc{SQuAD 2.0}}} \\
    \cmidrule{3-4}
    & & Overall & (Ans./Unans.) \\
    \midrule
    \texttt{distilbert-base} & $-0.6$ & $-0.71$ & $_{(-2.76/1.71)}$ \\
    \texttt{bert-base-cased} & $-0.21$ & $-0.63$ & $_{(-1.84/0.6)}$ \\
    \texttt{bert-base-uncased} & $-0.87$ & $-0.49$ & $_{(-1.88/0.94)}$ \\
    \texttt{bert-large-cased} & $-0.63$ & $-0.53$ & $_{(-1.61/0.55)}$ \\
    \texttt{bert-large-uncased} & $-0.35$ & $-1.38$ & $_{(-2.51/-0.24)}$ \\
    \texttt{spanbert-base-cased} & $-0.26$ & $-1.24$ & $_{(-2.66/0.15)}$ \\
    \texttt{spanbert-large-cased} & $-0.51$ & $-1.20$ & $_{(-1.9/-0.56)}$ \\
    \texttt{roberta-base} & $-0.61$ & $-0.60$ & $_{(-2.09/0.81)}$ \\
    \texttt{roberta-large} & $-0.29$ & $-1.52$ & $_{(-2.6/-0.54)}$ \\
    \texttt{albert-base-v1} & $-1.0$ & $-1.07$ & $_{(-2.02/-0.22)}$ \\
    \texttt{albert-base-v2} & $-0.34$ & $-1.08$ & $_{(-2.03/-0.22)}$ \\
    \texttt{albert-large-v1} & $-0.42$ & $-0.41$ & $_{(-1.42/0.52)}$ \\
    \texttt{albert-large-v2} & $-0.8$ & $-0.69$ & $_{(-1.66/0.22)}$ \\
    \texttt{albert-xxlarge-v1} & $-0.75$ & $-1.23$ & $_{(-3.06/0.49)}$ \\
    \texttt{albert-xxlarge-v2} & $-0.46$ & $-1.28$ & $_{(-3.02/0.36)}$ \\
    \texttt{deberta-large} & $-0.52$ & $-1.05$ & $_{(-2.2/0.0)}$ \\
    \bottomrule
    \end{tabular}
}
\caption{Relative F1 change (\%) for encoder-only MRC systems subjecting to natural perturbations. For \textsc{SQuAD 2.0}, the overall values are broken down to answerable and unanswerable questions, respectively.}
\label{tab:natural_encoder}
\end{table}

\subsection{Which Categories of Natural Perturbation Lead to Model Failure?}
\label{subsec:Error Analysis}

To investigate the sources of natural perturbation and reveal the perturbation phenomena contributing to encoder-only models' error, we manually label linguistic features between passages where models succeed and fail, to identify how they differ.

\begin{figure}[htb!]
    \centering
    \includegraphics[width=\columnwidth]{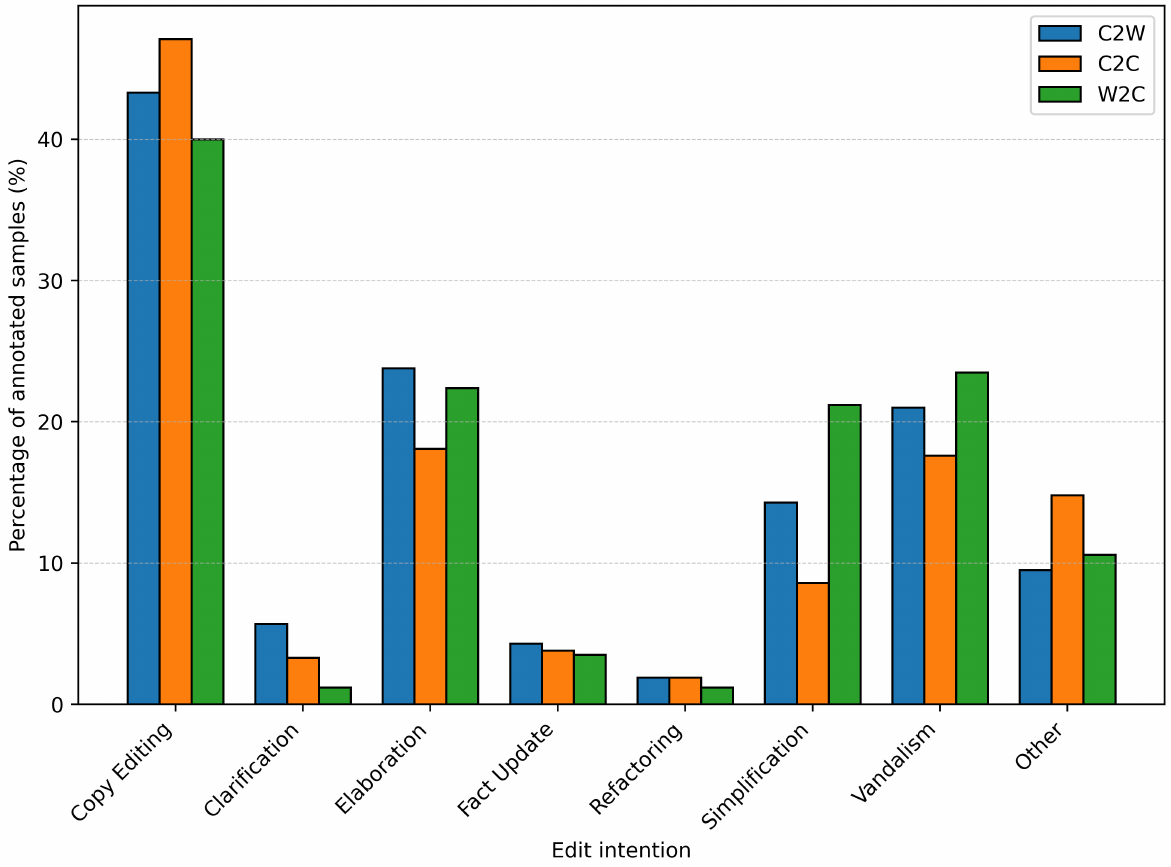}
    \caption{The percentage (\%) of samples annotated with each edit intention in the C2W, C2C and W2C categories. The percentages do not add up to $100$\% because a single revision may fall into multiple intentions.}
    \label{fig:error_analysis}
\end{figure}

Within the original and the naturally perturbed test set pair generated based on \textsc{SQuAD 2.0} development set, we first identify $384$ instances where at least one encoder-only model succeeds on the original but fails\footnote{For answerable questions, a model's prediction is considered correct if Exact Match (EM) score equals $1$, and incorrect if F1 score is $0$ or it determines the question is unanswerable. For unanswerable questions, a model's prediction is correct if it predicts the question is unanswerable, and wrong if it provides an answer span.} on the perturbed (i.e., being adversarial), and then randomly select the same number of instances on which all encoder-only models succeed on both the original and perturbed versions \citep{naik-etal-2018-stress}. We refer to these two types of instances as C2W (correct to wrong) and C2C (correct to correct) instances, respectively. Among the identified C2W and C2C instances, we further remove duplicates, resulting in $210$ and $244$ unique original and perturbed paragraph pairs, respectively. Furthermore, as natural perturbation can occasionally help the model to get the answer correct,
we also filter $85$ unique W2C (wrong to correct) instances on which at least two encoder-only models fail on the original but succeed on the perturbed. Finally, utilising an $8$-category taxonomy of the semantic edit intentions in Wikipedia revisions derived from~\citet{yang-etal-2017-identifying-semantic}, the chosen $210$ samples of C2W and C2C, as well as the $85$ W2C were annotated, with $20$\% of the annotated C2W and C2C examples presented to a second annotator for additional validation. See Appendix~\ref{sec:Human Annotation Instructions} for the instruction provided to the annotators, along with detailed explanations of each edit intention. We calculate the (micro-averaged) F1 score to evaluate the inter-annotator agreement, which is $0.82$. This suggests that the annotators' annotations align closely. Figure \ref{fig:error_analysis} reports the annotation results.

Distribution of perturbation types shown in Figure~\ref{fig:error_analysis} generally aligns with the edit intentions distribution annotated in \citep{yang-etal-2017-identifying-semantic}, with \textit{Copy Editing} and \textit{Elaboration} appearing more frequently than others, such as \textit{Clarification}, \textit{Fact Update}, and \textit{Refactoring}. This reflects the inherent characteristics of Wikipedia revisions. From Figure \ref{fig:error_analysis}, we observe that there is no significant difference in the distribution of annotated edit intentions between C2W and C2C examples, suggesting that 
\textbf{\textit{though these types of natural perturbations confuse the encoder-only MRC models, there seems no correlation with human-perceivable features
}}. 
A roughly similar distribution is also observed in the W2C examples, which indicates that these natural perturbation types can also facilitate correct answers by the models, i.e., being beneficial.
These demonstrate that on \textsc{SQuAD 2.0}, there might be no correlation between the quality of the naturally perturbed passage and its potential for being adversarial\footnote{We also find little or no significant correlation between the perturbation magnitude (measured as byte-level changes between the original and perturbed passages) and model failure, with point biserial correlation coefficient close to 0.}. Certain text edits aimed at improving the passage quality, such as \textit{Copy Editing} and \textit{Elaboration}, do render the perturbation adversarial, whereas edits intended to damage the article may not consistently result in adversarial instances; in fact, \textit{vandalism} can even assist models in providing correct answers. Instead, we infer that whether an edit to the passage can render the MRC instance adversarial or not depends on the location of the edits in relation to the question. Among the $384$ C2W and C2C examples, we measure the proportion of answerable questions with the answer sentence(s) in the original passage remaining unmodified in the naturally perturbed version, which is $34.5$\% and $71.5$\%, respectively. This confirms our hypothesis that if the edits affect the answer sentence(s), there is a higher likelihood of the perturbed example becoming adversarial; otherwise, it might not. \textit{Copy Editing} appears to alter the answer sentences in the reading passage more frequently, making it the most impactful category that confuses models (contributing to more than 40\% of error cases), while other types have a lesser effect. Appendix \ref{sec:Demonstration of Perturbed MRC Examples for Encoder-only Models} presents one perturbed example for each of the C2W, C2C, and W2C categories, respectively, along with the annotated natural perturbation type(s).

\subsection{To What Extent Do Natural Adversarial Examples Preserve Validity?}
\label{subsec:Validity of Nature Adversarial Examples}

To accurately assess a model's robustness under perturbation, it is vital to examine the validity of adversarial example, i.e., whether humans can still find the correct answer under the perturbation~\citep{dyrmishi-etal-2023-humans}. Two human annotators are recruited to verify the validity of the $210$ C2W examples in Section~\ref{subsec:Error Analysis} and the inter-annotator agreement is measured by computing the Cohen's $\kappa$ coefficient \citep{doi:10.1177/001316446002000104}. We then involve a third human annotator to annotate the adversarial examples on which the first two annotators disagree and take the majority label as ground truth. This validity verification process is detailed in Appendix \ref{sec:Process of Adversarial Validity Verification}. Out of 210 C2W examples, we find that $86$\% of them are valid ($0.77$ Cohen's $\kappa$), indicating that \textbf{\textit{a substantial proportion of natural adversarial examples for encoder-only MRC model(s) are valid}}.

\subsection{Can Errors from Encoder-only Models Affect Other Architectures?}
\label{subsec:Can Errors from Encoder-only Models Affect Other Architectures?}

We further investigate whether the errors identified in encoder-only models carry over to other more recent models and architectures, as SOTA advancements in NLP would suggest otherwise. Therefore, we propose an exhaustive search algorithm (Appendix \ref{sec:Exhaustive Search Algorithm for Challenging Test Set Construction}) to zoom in on the errors of encoder-only models as much as possible, curate the challenging natural perturbed test set, and finally examine the performance of Flan-T5 and LLMs. With the development set of \textsc{SQuAD 1.1} and \textsc{SQuAD 2.0} as the source, the algorithm results in two challenge perturbed test sets: \textsc{Nat\_V1\_Challenge} ($184$ contexts, $234$ questions) and \textsc{Nat\_V2\_Challenge} ($214$ contexts, $442$ questions ($226$ unanswerable)).
Table \ref{tab:natural_all} shows the evaluation results
on the newly generated challenge test sets. From the table, we observe that \textbf{\textit{the errors caused by natural perturbation in encoder-only MRC models transfer to both \texttt{Flan-T5} and LLMs}}. On the \textsc{Nat\_V1\_Challenge}, \texttt{flan-t5-small} demonstrates the greatest susceptibility to natural perturbations, experiencing a $14.27$\% decrease in F1; while among LLMs, \texttt{Gemma-7B-IT} emerges as the least robust, with a $16.66$\% IM drop, followed by \texttt{Gemma-2B-IT} ($-15.83$\%) and \texttt{Llama-3.1-8B-Instruct} ($-15.61$\%). Transitioning to the \textsc{Nat\_V2\_Challenge}, the base version of \texttt{flan-t5} exhibits the largest performance decline at $13.83$\% and \texttt{Falcon-7B-Instruct} stands out as the LLM with the lowest robustness ($-28.28$\%). Other LLMs such as \texttt{Qwen2.5-7B-Instruct} and \texttt{deepseek-llm-7b-chat} also show severe robustness loss, with drops of $12.21$\% and $11.29$\%, respectively. Further, we observe that the robustness of models under natural perturbations does not necessarily size-dependent. While larger models tend to exhibit greater robustness in some cases (e.g., \texttt{Qwen2.5-14B-Instruct} vs. \texttt{Qwen2.5-3B-Instruct}), exceptions within the \texttt{Falcon} and \texttt{Llama} model series suggest that factors beyond model size--such as training corpora, training and fine-tuning methodology, and architectural differences may also significantly affect susceptibility to natural perturbations.
In Appendix \ref{sec:Natural Adversarial Samples for LLMs}, we showcase two adversarial examples targeting LLMs sourced from our generated challenge sets.

\begin{table}[htb!]
    \centering
    \resizebox{\columnwidth}{!}{
    \begin{tabular}{lllll}
    \toprule
    \textbf{Model}&\multicolumn{4}{c}{\textbf{Performance}}\\\cline{3-4}&\multicolumn{4}{c}{\textbf{\textit{original} vs. \textit{perturbed}}}\\
    \cline{2-5}
    &\multicolumn{2}{c}{\textsc{Nat\_V1\_Challenge}}&\multicolumn{2}{c}{\textsc{Nat\_V2\_Challenge}}\\
    \midrule
\texttt{flan-t5-small}&$58.76$/$64.76$&$48.58$/$55.52$$_{-14.27}$&$42.57$/$44.57$&$39.71$/$41.81$$_{-6.19}$\\
\texttt{flan-t5-base}&$79.49$/$85.01$&$66.1$/$73.42$$_{-13.63}$&$70.66$/$72.85$&$61.16$/$62.78$$_{-13.83}$\\
\texttt{flan-t5-large}&$88.1$/$92.53$&$76.57$/$82.31$$_{-11.05}$&$79.11$/$81.01$&$70.14$/$72.13$$_{-10.96}$\\
\texttt{flan-t5-xl}&$86.25$/$91.57$&$75.0$/$81.45$$_{-11.05}$&$83.71$/$85.84$&$73.19$/$74.86$$_{-12.79}$\\
\texttt{GPT-3.5-turbo-0125}&$91.03$&$83.33$$_{-8.46}$&$51.58$&$47.06$$_{-8.76}$\\
\texttt{gpt-4o-2024-11-20}&$93.16$&$85.9$$_{-7.79}$&$80.09$&$75.11$$_{-6.22}$\\
\texttt{Gemma-2B-IT}&$51.28$&$43.16$$_{-15.83}$&$55.66$&$50.23$$_{-9.76}$\\
\texttt{Gemma-7B-IT}&$82.05$&$68.38$$_{-16.66}$&$59.95$&$57.01$$_{-4.9}$\\
\texttt{Gemma 2-2b-IT}&$85.47$&$78.21$$_{-8.49}$&$48.87$&$43.44$$_{-11.11}$\\
\texttt{Gemma 2-9b-IT}&$89.32$&$81.62$$_{-8.62}$&$64.93$&$59.95$$_{-7.67}$\\
\texttt{Llama 2-chat-7B}&$82.91$&$73.93$$_{-10.83}$&$41.63$&$38.69$$_{-7.06}$\\
\texttt{Llama 2-chat-13B}&$80.77$&$73.93$$_{-8.47}$&$46.83$&$41.18$$_{-12.06}$\\
\texttt{Llama-3-8B-Instruct}&$88.89$&$77.35$$_{-12.98}$&$51.81$&$46.61$$_{-10.04}$\\
\texttt{Llama-3.1-8B-Instruct}&$87.61$&$73.93$$_{-15.61}$&$61.31$&$55.43$$_{-9.59}$\\
\texttt{Llama-3.2-1B-Instruct}&$54.27$&$47.86$$_{-11.81}$&$35.29$&$32.13$$_{-8.95}$\\
\texttt{Llama-3.2-3B-Instruct}&$81.2$&$71.37$$_{-12.11}$&$48.42$&$43.44$$_{-10.29}$\\
\texttt{Mistral-7B-Instruct-v0.2}&$84.19$&$73.08$$_{-13.2}$&$54.98$&$51.36$$_{-6.58}$\\
\texttt{OLMo-7B-0724-Instruct}&$90.17$&$82.91$$_{-8.05}$&$51.36$&$49.1$$_{-4.4}$\\
\texttt{Qwen2.5-3B-Instruct}&$78.63$&$68.38$$_{-13.04}$&$61.31$&$54.07$$_{-11.81}$\\
\texttt{Qwen2.5-7B-Instruct}&$88.03$&$81.2$$_{-7.76}$&$76.02$&$66.74$$_{-12.21}$\\
\texttt{Qwen2.5-14B-Instruct}&$92.31$&$81.62$$_{-11.58}$&$80.54$&$74.21$$_{-7.86}$\\
\texttt{Falcon-7B-Instruct}&$53.42$&$50.00$$_{-6.4}$&$32.81$&$23.53$$_{-28.28}$\\
\texttt{Falcon-40B-Instruct}&$69.66$&$62.82$$_{-9.82}$&$38.69$&$36.88$$_{-4.68}$\\
\texttt{Falcon3-7B-Instruct}&$88.03$&$79.49$$_{-9.7}$&$59.28$&$55.43$$_{-6.49}$\\
\texttt{Falcon3-10B-Instruct}&$90.6$&$82.91$$_{-8.49}$&$64.48$&$59.73$$_{-7.37}$\\
\texttt{deepseek-llm-7b-chat}&$70.51$&$64.1$$_{-9.09}$&$42.08$&$37.33$$_{-11.29}$\\
\bottomrule
    \end{tabular}
    }
    \caption{Performance (\%) of Flan-T5 and SOTA LLMs on~\textsc{Nat\_V1\_Challenge} and~\textsc{Nat\_V2\_Challenge}. Values in smaller font are changes (\%) relative to the original performance of the model.}
    \label{tab:natural_all}
\end{table}

\subsection{Do Our Findings Generalise to Other Challenging MRC Datasets?}
\label{subsec:Do Our Findings Generalise to Other MRC Datasets?}

The two \textsc{SQuAD} datasets investigated previously are relatively simple, as they lack challenging features~\citep{schlegel-etal-2020-framework}, leading to super-human performance of MRC models~\citep{Lan2020ALBERT:}. To generalise our findings to more challenging MRC benchmarks, we apply the natural perturbation methodology (Section \ref{sec:Natural Perturbation Pipeline}) to the development set of four more datasets and assess the performance changes of multiple LLMs, as shown in Table~\ref{tab:natural_others}. For \textsc{DROP}~\citep{dua-etal-2019-drop}, we first use the \texttt{GPT-4o mini} to infer the likely Wikipedia article title from which each passage is retrieved~\footnote{This is because the raw Wikipedia title information cannot be found in the original development set of~\textsc{DROP}. We use the prompt: “Given a reading paragraph, return the Wikipedia page title from which it is likely retrieved.”} and extract
the revision histories for those articles. For \textsc{HotpotQA}~\citep{yang-etal-2018-hotpotqa}, we only perturb the paragraphs containing the supporting facts, while the distracting passages remain unchanged. Furthermore, using the validity verification method described in Section~\ref{subsec:Validity of Nature Adversarial Examples}, we manually verify the validity percentage of all adversarial examples in~\textsc{DROP} and~\textsc{TyDi QA}, as well as $50$ randomly selected adversarial examples from~\textsc{BoolQ} and~\textsc{HotpotQA}, as reported in Table~\ref{tab:natural_others}.

\begin{table}[htb!]
    \centering
    \resizebox{\columnwidth}{!}{
    \begin{tabular}{lcccc}
        \toprule
        \textbf{LLM}&\multicolumn{4}{c}{\textbf{IM Relative Change (\%)}}\\
        \cmidrule{2-5}
        &\textsc{BoolQ}&\textsc{DROP}&\textsc{HotpotQA}&\textsc{TYDI QA}\\
        \midrule
        \textit{adversarial validity (\%) (Cohen's $\kappa$)}&$72$ (0.54)&$85.7$ (0.46)&$88$ (0.6)&$87.5$ (0.52)\\
        \midrule
        \texttt{Gemma 2-2b-IT}&$-3.91$&$-2.22$&$-$&$-1.61$\\
        \texttt{Gemma 2-9b-IT}&$-3.92$&$-1.69$&$-2.21$&$-1.51$\\
        \texttt{Llama-3.1-8B-Instruct}&$-3.05$&$-7.13$&$-0.91$&$3.17$\\
        \texttt{Llama-3.2-1B-Instruct}&$-3.81$&$-9.98$&$-1.73$&$-9.1$\\
        \texttt{Llama-3.2-3B-Instruct}&$-3.74$&$-$&$-2.05$&$-$\\
        \texttt{Mistral-7B-Instruct-v0.2}&$-5.25$&$-1.85$&$-1.16$&$-$\\
        \texttt{OLMo-7B-0724-Instruct}&$-4.49$&$-7.9$&$-2.36$&$2.94$\\
        \texttt{Qwen2.5-7B-Instruct}&$-4.24$&$-3.85$&$-1.18$&$-2.78$\\
        \texttt{Qwen2.5-14B-Instruct}&$-3.22$&$-$&$-1.84$&$1.38$\\
        \texttt{Falcon3-7B-Instruct}&$-5.1$&$2.04$&$-0.06$&$-8.82$\\
        \texttt{Falcon3-10B-Instruct}&$-3.58$&$1.8$&$-1.81$&$-4.22$\\
        \bottomrule
    \end{tabular}}
    \caption{IM changes (\%) of SOTA LLMs on naturally perturbed test set of other more challenging MRC datasets.}
    \label{tab:natural_others}
\end{table}

Overall, \textit{\textbf{when natural perturbations are applied to other more challenging benchmarks, SOTA LLMs also exhibit a lack of robustness}}, with the largest $9.98$ performance decrease observed for~\texttt{Llama-3.2-1B-Instruct} on~\textsc{DROP}. This further demonstrates the broad and severe impact of natural perturbations on diverse MRC tasks, especially in light of on three out of four benchmarks, our human annotators are still able to correctly answer over $85$\% of the adversarial examples. \textsc{BoolQ} exhibits a lower adversarial validity rate ($72$\%). However, in most cases, this is not due to the perturbation degrading the passage, but rather the poor quality of the original instance~\citep{tedeschi-etal-2023-whats}, i.e., even with the corresponding original passage, humans are unable to assign the correct label. For instance, from the annotators' perspective, the question itself may be ambiguous, such as \textit{``do you need a visa to visit oman''}, or entirely unanswerable due to missing information in the original passage. Our human annotations also observe diverse error patterns in LLMs caused by natural perturbations (e.g., \textit{Copy Editing}, \textit{Elaboration} and \textit{Vandalism}), as what are presented in Figure~\ref{fig:error_analysis}. Finally, we note that for a very small proportion of adversarial examples, the limitations of the IM metric mean that the model’s answer under perturbation does not necessarily indicate an error, emphasising the need for future efforts to address this concern.

\section{Dealing With Natural Perturbations}
\label{sec:Dealing With Natural Perturbations}

In this section, we provide an initial exploration of methods to defend against natural perturbations.
To enhance encoder-only model robustness, we first conduct adversarial training by identifying six encoder-only model architectures that already exhibit the highest robustness to natural perturbations in their respective categories (except \texttt{albert-xxlarge-v2} on \textsc{Nat\_V2\_Challenge}), and presenting them with both original training data and the generated naturally perturbed training examples. We extract the entire Wikipedia revision histories for the $392$ articles in the original \textsc{SQuAD} training set, and then obtain $5,262$ (with $22,033$ questions) and $5,311$ (with $32,993$ questions) perturbed contexts to augment the original \textsc{SQuAD 1.1} and \textsc{SQuAD 2.0} training set, respectively, using the methodology described in Section~\ref{sec:Natural Perturbation Pipeline}. Table~\ref{tab:nat_retrain} compares the performance of these models on \textsc{Nat\_V1\_Challenge} and \textsc{Nat\_V2\_Challenge}, before and after retraining.


\begin{table}[htb!]
    \centering
    \resizebox{\columnwidth}{!}{
    \begin{tabular}{lllll}
    \toprule
    \textbf{Model}&\multicolumn{4}{c}{\textbf{Performance}}\\&\multicolumn{4}{c}{\textbf{(EM/F1)}}\\\cline{3-4}&\multicolumn{4}{c}{\textbf{\textit{original} vs. \textit{perturbed}}}\\
    \cline{2-5}
    &\multicolumn{2}{c}{\textsc{Nat\_V1\_Challenge}}&\multicolumn{2}{c}{\textsc{Nat\_V2\_Challenge}}\\
    \midrule
    \texttt{distilbert-base}&$64.53$/$70.45$&$41.03$/$47.6$$_{-32.43}$&$56.56$/$59.08$&$41.18$/$43.3$$_{-26.71}$\\
    \rowcolor{lightgray} \cellcolor{white}&$57.26$/$63.44$&$43.59$/$51.87$$_{-18.24}$&$53.17$/$55.4$&$43.89$/$45.51$$_{-17.85}$\\
    \texttt{bert-large-cased}&$79.06$/$83.66$&$63.68$/$70.23$$_{-16.05}$&$66.29$/$68.35$&$53.17$/$55.04$$_{-19.47}$\\
    \rowcolor{lightgray} \cellcolor{white}&$74.79$/$80.14$&$59.83$/$67.5$$_{-15.77}$&$67.87$/$69.31$&$58.37$/$59.53$$_{-14.11}$\\
    \texttt{spanbert-large-cased}&$84.19$/$88.2$&$67.95$/$74.77$$_{-15.23}$&$78.73$/$80.68$&$62.44$/$64.99$$_{-19.45}$\\
    \rowcolor{lightgray} \cellcolor{white}&$82.48$/$86.6$&$69.66$/$76.05$$_{-12.18}$&$78.28$/$80.0$&$65.61$/$67.12$$_{-16.1}$\\
    \texttt{roberta-large}&$86.75$/$90.21$&$73.93$/$79.47$$_{-11.91}$&$82.13$/$84.27$&$66.29$/$68.52$$_{-18.69}$\\
    \rowcolor{lightgray} \cellcolor{white}&$83.33$/$87.15$&$70.94$/$76.53$$_{-12.19}$&$81.22$/$82.67$&$70.59$/$71.84$$_{-13.1}$\\
    \texttt{albert-xxlarge-v2}&$84.62$/$89.64$&$73.93$/$78.77$$_{-12.13}$&$84.62$/$86.07$&$68.1$/$69.61$$_{-19.12}$\\
    \rowcolor{lightgray} \cellcolor{white}&$86.32$/$90.93$&$75.64$/$81.07$$_{-10.84}$&$82.58$/$84.08$&$70.59$/$72.78$$_{-13.44}$\\
    \texttt{deberta-large}&$88.46$/$92.5$&$73.5$/$78.48$$_{-15.16}$&$85.07$/$86.65$&$71.49$/$73.0$$_{-15.75}$\\
    \rowcolor{lightgray} \cellcolor{white}&$88.03$/$91.84$&$76.92$/$81.53$$_{-11.23}$&$83.03$/$85.1$&$72.62$/$74.48$$_{-12.48}$\\
    \bottomrule
    \end{tabular}}
    \caption{Comparison of the performance of several encoder-only MRC systems on \textsc{Nat\_V1\_Challenge} and \textsc{Nat\_V2\_Challenge}, before and after re-training. The results shown in the shaded areas represent the performance of the model retrained on the augmented training set with naturally perturbed instances.}
    \label{tab:nat_retrain}
\end{table}

\begin{figure}[htb!]
    \centering
    \includegraphics[width=\columnwidth]{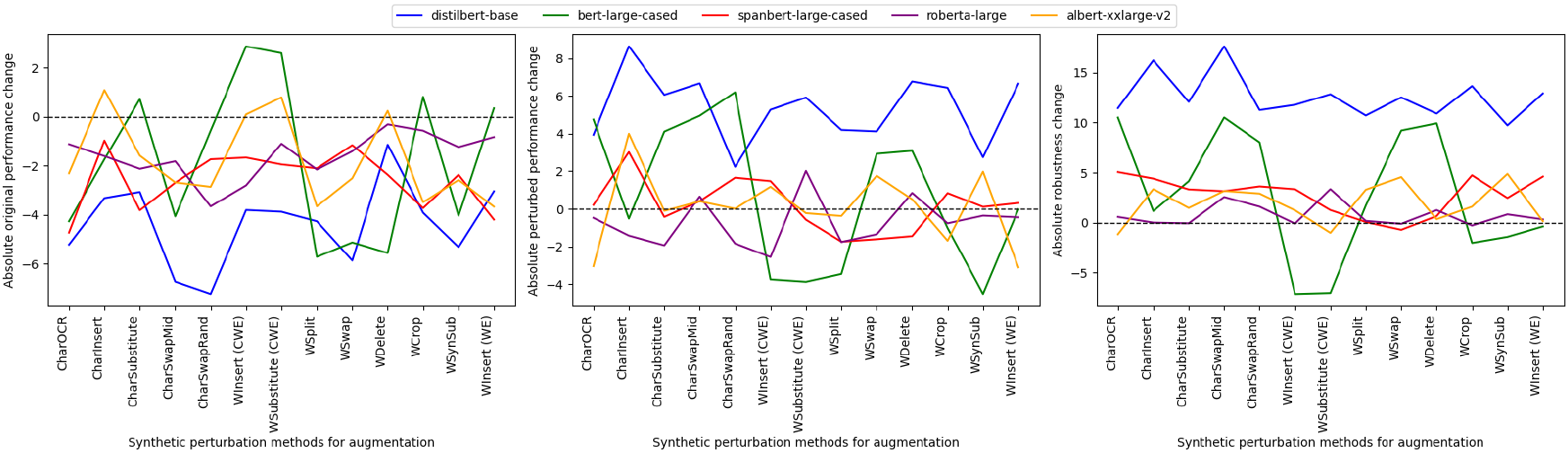}
    \caption{Absolute changes in original and perturbed performance (F1), as well as the robustness of five encoder-only models under natural perturbations (on \textsc{Nat\_V1\_Challenge}), following retraining with each synthetic perturbation.}
    \label{fig:syn_helps_nat_others}
\end{figure}

Apart from re-training with the same type of noise, we also ask whether exposing models to synthetic perturbations can help them confront natural ones. Therefore, we incorporate thirteen synthetic perturbation techniques spanning character and word levels (see Appendix \ref{sec:Synthetic Perturbation Methods}). Afterwards, we first retrain \texttt{deberta-large} with perturbed training samples generated by each synthetic perturbation method, respectively, and assess the performance changes compared to the vanilla version on both \textsc{Nat\_V1\_Challenge} and \textsc{Nat\_V2\_Challenge} (Figure \ref{fig:syn_helps_nat} in Appendix \ref{sec:Impact of Synthetic Adversarial Training}). 
As we observe that synthetic adversarial training can assist \texttt{deberta-large} in handling natural perturbations, we further retrain five other models in the same manner
and quantify the performance difference on \textsc{Nat\_V1\_Challenge} compared to the vanilla version, as shown in Figure \ref{fig:syn_helps_nat_others}.

In general, for encoder-only MRC models, retraining with natural perturbations enhances the performance on naturally perturbed test sets and improves the robustness to such perturbations as well, though this can lead to varying reductions in performance on the clean test set. Encouragingly, adversarial training with synthetically perturbed examples benefits the model's capability to handle natural perturbations as well, a phenomenon differs from what is reported in machine translation task \citep{belinkov2018synthetic}. In some cases, the improvement even exceeds what achieved by retraining the model on natural perturbations alone. We also observe that the effectiveness of adversarial training varies with model size and architecture. Generally, adversarial training brings the most significant benefits for the weakest \texttt{distilbert-base}, with the benefits diminishing in larger and more complex model architectures.

Similarly, for the LLMs, we adopt a few-shot prompting approach by including both the original MRC instance and its naturally or synthetically perturbed counterpart as demonstrations, and assess how model performance and robustness change compared to the zero-shot setting (see Table~\ref{tab:rob_improve_nature} and Table~\ref{tab:rob_improve_synthetic}). A total of two original–perturbed instance pairs are used, with the original samples taken from the~\textsc{SQuAD} training set. Although not widely observed, in certain cases, in-context demonstrations can improve an LLM's resilience to natural perturbations, regardless of whether natural or synthetic perturbed examples are demonstrated. This phenomenon is particularly evident in models such as~\texttt{Llama-3.2-3B-Instruct}, \texttt{OLMo-7B-0724-Instruct} and \texttt{Falcon3-10B-Instruct}. However, it can also have detrimental effects, further decreasing LLM robustness and resulting in a performance decline on both the clean and naturally perturbed test sets.

\begin{table}[htb!]
    \centering
    \resizebox{\columnwidth}{!}{
    \begin{tabular}{lllllll}
    \toprule
    \textbf{LLM}&\multicolumn{3}{c}{\textsc{Nat\_V1\_Challenge}}&\multicolumn{3}{c}{\textsc{Nat\_V2\_Challenge}}\\
    \cline{2-7}
    & \textit{Orig.$/$Pert.} & \textit{IM Drop} & \textit{zero-shot} & \textit{Orig.$/$Pert.} & \textit{IM Drop} & \textit{zero-shot} \\
    \midrule
    \texttt{Gemma 2-2b-IT}&$80.77$/$72.22$ $\downarrow$&$-10.59$&$-8.49$&$45.93$/$42.53$ $\downarrow$&\underline{$-7.4$}&$-11.11$\\
    \texttt{Gemma 2-9b-IT}&$\uparrow$ $91.88$/$79.91$ $\downarrow$&$-13.03$&$-8.62$&$62.9$/$57.24$ $\downarrow$&$-9.0$&$-7.67$\\
    \texttt{Llama-3.1-8B-Instruct}&$\downarrow$ $85.04$/$74.79$ $\uparrow$&\underline{$-12.05$}&$-15.61$&$45.25$/$41.86$ $\downarrow$&\underline{$-7.49$}&$-9.59$\\
    \texttt{Llama-3.2-3B-Instruct}&$70.09$/$62.82$ $\downarrow$&\underline{$-10.37$}&$-12.11$&$43.67$/$43.44$ $\downarrow$&\underline{$-0.53$}&$-10.29$\\
    \texttt{Mistral-7B-Instruct-v0.2}&$\downarrow$ $83.33$/$77.78$ $\uparrow$&\underline{$-6.66$}&$-13.2$&$50.45$/$46.61$ $\downarrow$&$-7.61$&$-6.58$\\
    \texttt{OLMo-7B-0724-Instruct}&$76.5$/$76.07$ $\downarrow$&\underline{$-0.56$}&$-8.05$&$53.62$/$51.58$ $\uparrow$&\underline{$-3.8$}&$-4.4$\\
    \texttt{Qwen2.5-3B-Instruct}&$61.11$/$47.44$ $\downarrow$&$-22.37$&$-13.04$&$67.42$/$61.54$ $\uparrow$&\underline{$-8.72$}&$-11.81$\\
    \texttt{Qwen2.5-7B-Instruct}&$86.32$/$73.08$ $\downarrow$&$-15.34$&$-7.76$&$76.24$/$69.46$ $\uparrow$&\underline{$-8.89$}&$-12.21$\\
    \texttt{Qwen2.5-14B-Instruct}&$85.47$/$72.65$ $\downarrow$&$-15.0$&$-11.58$&$80.09$/$73.76$ $\downarrow$&$-7.9$&$-7.86$\\
    \texttt{Falcon3-7B-Instruct}&$86.32$/$75.21$ $\downarrow$&$-12.87$&$-9.7$&$\uparrow$ $60.86$/$54.98$ $\downarrow$&$-9.66$&$-6.49$\\
    \texttt{Falcon3-10B-Instruct}&$85.9$/$79.06$ $\downarrow$&\underline{$-7.96$}&$-8.49$&$61.54$/$59.5$ $\downarrow$&\underline{$-3.31$}&$-7.37$\\
    \texttt{deepseek-llm-7b-chat}&$\uparrow$ $70.94$/$55.98$ $\downarrow$&$-21.09$&$-9.09$&$59.73$/$51.58$ $\uparrow$&$-13.64$&$-11.29$\\
    \bottomrule
    \end{tabular}}
    \caption{Performance and IM drop of LLMs in the few-shot setting with both original and \textit{naturally} perturbed MRC instances demonstrated. \textit{zero-shot} represents the IM drop in the zero-shot setting, adopted from Table~\ref{tab:natural_all}. Results that evidence robustness improvement in the few-shot setting are underlined.}
    \label{tab:rob_improve_nature}
\end{table}

\begin{table}[htb!]
    \centering
    \resizebox{\columnwidth}{!}{
    \begin{tabular}{lllllll}
    \toprule
    \textbf{LLM}&\multicolumn{3}{c}{\textsc{Nat\_V1\_Challenge}}&\multicolumn{3}{c}{\textsc{Nat\_V2\_Challenge}}\\
    \cline{2-7}
    & \textit{Orig.$/$Pert.} & \textit{IM Drop} & \textit{zero-shot} & \textit{Orig.$/$Pert.} & \textit{IM Drop} & \textit{zero-shot} \\
    \midrule
    \texttt{Gemma 2-2b-IT}&$82.48$/$72.22$ $\downarrow$&$-12.44$&$-8.49$&$47.29$/$42.53$ $\downarrow$&\underline{$-10.07$}&$-11.11$\\
    \texttt{Gemma 2-9b-IT}&$\uparrow$ $91.45$/$81.62$&$-10.75$&$-8.62$&$61.54$/$56.33$ $\downarrow$&$-8.47$&$-7.67$\\
    \texttt{Llama-3.1-8B-Instruct}&$83.33$/$72.65$ $\downarrow$&\underline{$-12.82$}&$-15.61$&$49.32$/$44.57$ $\downarrow$&$-9.63$&$-9.59$\\
    \texttt{Llama-3.2-3B-Instruct}&$66.24$/$62.82$ $\downarrow$&\underline{$-5.16$}&$-12.11$&$42.53$/$41.86$ $\downarrow$&\underline{$-1.58$}&$-10.29$\\
    \texttt{Mistral-7B-Instruct-v0.2}&$\downarrow$ $81.62$/$76.92$ $\uparrow$&\underline{$-5.76$}&$-13.2$&$50.68$/$46.61$ $\downarrow$&$-8.03$&$-6.58$\\
    \texttt{OLMo-7B-0724-Instruct}&$73.84$/$75.53$ $\downarrow$&\underline{$2.29$}&$-8.05$&$54.98$/$53.39$ $\uparrow$&\underline{$-2.89$}&$-4.4$\\
    \texttt{Qwen2.5-3B-Instruct}&$59.4$/$51.71$ $\downarrow$&\underline{$-12.95$}&$-13.04$&$66.52$/$61.09$ $\uparrow$&\underline{$-8.16$}&$-11.81$\\
    \texttt{Qwen2.5-7B-Instruct}&$85.04$/$73.5$ $\downarrow$&$-13.57$&$-7.76$&$76.47$/$69.0$ $\uparrow$&\underline{$-9.77$}&$-12.21$\\
    \texttt{Qwen2.5-14B-Instruct}&$84.19$/$73.5$ $\downarrow$&$-12.7$&$-11.58$&$\uparrow$ $81.9$/$74.21$&$-9.39$&$-7.86$\\
    \texttt{Falcon3-7B-Instruct}&$85.47$/$74.36$ $\downarrow$&$-13.0$&$-9.7$&$61.09$/$56.56$ $\uparrow$&$-7.42$&$-6.49$\\
    \texttt{Falcon3-10B-Instruct}&$85.9$/$79.91$ $\downarrow$&\underline{$-6.97$}&$-8.49$&$61.31$/$57.92$ $\downarrow$&\underline{$-5.53$}&$-7.37$\\
    \texttt{deepseek-llm-7b-chat}&$\uparrow$ $72.22$/$56.41$ $\downarrow$&$-21.89$&$-9.09$&$61.09$/$53.62$ $\uparrow$&$-12.23$&$-11.29$\\
    \bottomrule
    \end{tabular}}
    \caption{Performance and IM drop of LLMs in the few-shot setting with both original and \textit{synthetically} perturbed MRC instances demonstrated. \textit{zero-shot} represents the IM drop in the zero-shot setting, adopted from Table~\ref{tab:natural_all}. Results that evidence robustness improvement in the few-shot setting are underlined.}
    \label{tab:rob_improve_synthetic}
\end{table}

\section{Conclusion}
\label{sec:Conclusion}

In this paper, we first study the robustness of MRC models to \textit{natural} perturbations, which occur under real-world conditions without intentional human intervention. Using the proposed evaluation framework, we show that certain naturally perturbed examples can indeed be adversarial, i.e., lead to model failure, even when the modifications aim to improve the overall passage quality. Natural perturbations also appear to differ significantly from synthetic ones, exhibiting a wide range of rich linguistic phenomena and may be more effective in generating valid adversarial instances. Adversarial training via augmentation with either naturally or synthetically perturbed samples is generally beneficial for enhancing the model's robustness to
natural perturbations; yet, it can decrease performance on clean test set. Future work includes the exploration of alternative natural perturbation approaches and the design of more effective defensive strategies.
\section*{Limitations}

We acknowledge several limitations in our work: (1) Our perturbation framework constructs natural perturbations from Wikipedia edit history and therefore only works with Wikipedia-based benchmarks. Since the phenomenon of natural perturbations is by no means limited to Wikipedia and can occur in any kind of text that evolves over time, future work should explore alternative methods to generate natural perturbations for non-Wikipedia MRC datasets. (2) As training data augmentation and in-context demonstration have a relatively limited impact, further research is needed to develop better techniques for improving the robustness of both encoder-only models and LLMs to natural perturbations, and to investigate the relationship between robustness to natural and synthetic perturbations. (3) Potential benchmark contamination may affect our findings on LLM evaluation. Investigating its extent and impact on LLM performance and robustness evaluation will be a focus of our future research efforts.

\section*{Ethical Considerations}

All datasets, extracted natural perturbations, and models used in this work are publicly available, used consistently with their intended purpose and under the permitted license. A very small proportion of natural perturbations may contain offensive content, as they come from reverted Wikipedia revisions intended to damage the articles. We include these to raise awareness within the community about their potential impact on MRC models and to call for methods to improve the safety of MRC models--especially those LLMs operating under such adversarial conditions. While our ultimate goal is to enhance model robustness, the findings from this work may carry the risk of being misused by malicious attackers to refine adversarial attack strategies and craft attacks against similar systems. Before starting the annotation task, we provide all annotators with clear instructions and inform the intended use of their annotations, obtaining their explicit consent. No private or sensitive information was collected, other than their annotations.

\bibliography{anthology,custom}

\appendix

\section{Encoder-only Model Parameters and Hyperparameters for Fine-tuning}
\label{sec:Encoder-only Model Parameters and Hyperparameters for Fine-tuning}

Table \ref{tab:eompahff} shows the hyperparameters used to fine-tune the pre-trained encoder-only MRC models in this work and their number of parameters contained.

\begin{table}[htb!]
    \centering
    \begin{tabular}{ccccc}
    \toprule
    Model$_{Parameters (M)}$&d&b&lr&ep\\
    \midrule
    \texttt{DistilBERT}$_{(66)}$&$384$&$8$&$3e-5$&$3$\\
    \texttt{BERT}$_{(110/340)}$&$384$&$8$&$3e-5$&$2$\\
    \texttt{SpanBERT}$_{(110/340)}$&$512$&$4$&$2e-5$&$4$\\
    \texttt{RoBERTa}$_{(125/355)}$&$384$&$8$&$3e-5$&$2$\\
    \texttt{ALBERT}$_{(11/17/223)}$&$384$&$4$&$3e-5$&$2$\\
    \texttt{DeBERTa}$_{(350)}$&$384$&$4$&$3e-6$&$3$\\
    \bottomrule
    \end{tabular}
    \caption{Number of parameters in each type of pre-trained encoder-only MRC model and the hyperparameters used to fine-tune them. For \texttt{BERT}, \texttt{SpanBERT}, \texttt{RoBERTa} and \texttt{ALBERT}, we show the number of model parameters in the order of \emph{base}, \emph{large} and \emph{xxlarge} (if applicable) version. d is the size of the token sequence fed into the model, b is the training batch size, lr is the learning rate, and ep is the number of training epochs. We used stride = $128$ for documents longer than d tokens.}
    \label{tab:eompahff}
\end{table}

\section{Instruction Templates for Flan-T5 Evaluation}
\label{sec:Instruction Templates for Flan-T5 Evaluation}

In Table \ref{tab:itffe}, we present the instruction templates employed in constructing the inputs to the \texttt{Flan-T5} model for the \textsc{SQuAD 1.1} format and \textsc{SQuAD 2.0} format test sets, respectively.

\begin{table}[htb!]
    \centering
    \begin{tabularx}{1\columnwidth}{p{0.05cm}p{7cm}}
    \hline
    \multicolumn{2}{c}{\textit{\textsc{SQuAD 1.1}}}\\
    \hline
    1&``Read this and answer the question\texttt{\string\n\string\n}\{context\}\texttt{\string\n\string\n}\{question\}''\\
    2&``\{context\}\texttt{\string\n}\{question\}''\\
    3&``Answer a question about this article:\texttt{\string\n}\{context\}\texttt{\string\n}\{question\}''\\
    4&``Here is a question about this article: \{context\}\texttt{\string\n}What is the answer to this question: \{question\}''\\
    5&``Article: \{context\}\texttt{\string\n\string\n}Question: \{question\}''\\
    6&``Article: \{context\}\texttt{\string\n\string\n}Now answer this question: \{question\}''\\
    \hline\hline
    \multicolumn{2}{c}{\textit{\textsc{SQuAD 2.0}}}\\
    \hline
    1&``Read this and answer the question. If the question is unanswerable, say \textbackslash``unanswerable\textbackslash".\texttt{\string\n\string\n}\{context\}\texttt{\string\n\string\n}\{question\}''\\
    2&``\{context\}\texttt{\string\n}\{question\} (If the question is unanswerable, say \textbackslash``unanswerable\textbackslash")''\\
    3&``\{context\}\texttt{\string\n}Try to answer this question if possible (otherwise reply \textbackslash``unanswerable\textbackslash"): \{question\}''\\
    4&``\{context\}\texttt{\string\n}If it is possible to answer this question, answer it for me (else, reply \textbackslash``unanswerable\textbackslash"): \{question\}''\\
    5&``\{context\}\texttt{\string\n\string\n}Answer this question, if possible (if impossible, reply \textbackslash``unanswerable\textbackslash"): \{question\}''\\
    6&``Read this: \{context\}\texttt{\string\n}Now answer this question, if there is an answer (If it cannot be answered, return \textbackslash``unanswerable\textbackslash"): \{question\}''\\
    \hline
    \end{tabularx}
    \caption{Various instruction templates for \texttt{Flan-T5} model evaluation.}
    \label{tab:itffe}
\end{table}

\section{MRC Prompts}
\label{sec:MRC Prompts}

We use the following zero-shot prompts to instruct the decoder-only models to generate responses in the task of MRC.

\textbf{\textsc{SQuAD 1.1}} \& \textbf{\textsc{TyDi QA}}:
\emph{Use the provided article delimited by triple quotes to answer question. Provide only the shortest continuous span from the context without any additional explanation.\texttt{\string\n\string\n}``````\{context\}"""\texttt{\string\n\string\n}Question: \{question\}}

\textbf{\textsc{SQuAD 2.0}}: \emph{Use the provided article delimited by triple quotes to answer question. Provide only the shortest continuous span from the context without any additional explanation. If the question is unanswerable, return ``unanswerable".\texttt{\string\n\string\n}``````\{context\}"""\texttt{\string\n\string\n}Question: \{question\}}

\textbf{\textsc{DROP} \& \textsc{HotpotQA}}: \emph{Use the provided article delimited by triple quotes to answer question. Provide only the answer without any additional explanation.\texttt{\string\n\string\n}``````\{context\}"""\texttt{\string\n\string\n}Question: \{question\}}

\textbf{\textsc{BoolQ}}: \emph{Use the provided article delimited by triple quotes to answer question. Return only TRUE or FALSE.\texttt{\string\n\string\n}``````\{context\}"""\texttt{\string\n\string\n}Question: \{question\}}

\section{Indicators of Unanswerable}
\label{sec:Indicators of Unanswerable}

We manually identify a set of phrases contained in the output of LLMs that indicate the unanswerability of the question, including \textit{``I cannot answer this/the question'', ``unanswerable'', ``There is no indication in the provided article'', ``The context provided does not provide enough information'', ``There is no reference in the given article'', ``The answer to the question is not provided in the given article'', ``it is not possible'', ``question cannot be answered''} and \textit{``context/question/article/text/article provided/passage does not''}.

\section{Impact of the Complete Set of Perturbed Instances on Encoder-Decoder and Decoder-Only Architectures}
\label{sec:Impact of the Complete Set of Perturbed Instances on Encoder-Decoder and Decoder-Only Architectures}

We supplement Table \ref{tab:natural_encoder} in Section \ref{subsec:Are Encoder-only MRC Models Resilient to Natural Perturbation?} with additional experiments on \texttt{Flan-T5} and some SOTA LLMs such as \texttt{Gemma 2}~\citep{Riviere2024Gemma2I} and \texttt{Llama 3.2}, to study the effect of all perturbed instances on these two architecture types (in addition to the encoder-only one). The results are presented in Table~\ref{tab:se}. From Table~\ref{tab:se}, we can see that similar to encoder-only models, \texttt{Flan-T5} and LLMs generally exhibit varying degrees of performance degradation under natural perturbations, but also exhibit considerable robustness.

\begin{table}[htb!]
\centering
\resizebox{\columnwidth}{!}{
\begin{tabular}{lcc}
\toprule
\textbf{Victim} & \textbf{\textsc{SQuAD 1.1}} & \textbf{\textsc{SQuAD 2.0}} \\ \midrule
\texttt{flan-t5-small} & $-0.69$ & $-0.64$ \\
\texttt{flan-t5-base} & $-0.91$ & $-1.32$ \\
\texttt{flan-t5-large} & $-0.77$ & $-1.13$ \\
\texttt{flan-t5-xl} & $-0.98$ & $-1.37$ \\
\texttt{Gemma 2-2b-IT} & $-$ & $-0.76$ \\
\texttt{Gemma 2-9b-IT} & $-0.89$ & $-0.92$ \\
\texttt{Llama-3.1-8B-Instruct} & $-0.38$ & $0.39$ \\
\texttt{Llama-3.2-3B-Instruct} & $-0.96$ & $-0.37$ \\
\texttt{Mistral-7B-Instruct-v0.2} & $0.39$ & $-1.28$ \\
\texttt{Falcon-7B-Instruct} & $-0.88$ & $-5.38$ \\
\texttt{Falcon-40B-Instruct} & $-0.80$ & $-$ \\
\bottomrule
\end{tabular}}
\caption{Performance change (\%) for \texttt{Flan-T5} and LLMs subjecting to natural perturbations.}
\label{tab:se}
\end{table}

We then measure the transferability of adversarial examples across all the evaluated model architectures and observe that these models share similar error patterns, with LLMs (especially \texttt{Falcon}) showing moderate differences. However, the lowest transferability metric is still as high as $0.86$.

\section{Human Annotation Instructions}
\label{sec:Human Annotation Instructions}

In Figure \ref{fig:hai}, we show the instructions given to human annotators for error analysis (Section \ref{subsec:Error Analysis}) and adversarial validity checking (Section \ref{subsec:Validity of Nature Adversarial Examples}), respectively.
All our human annotators are university students in the United Kingdom and China. Before commencing each task, we ask the annotators to annotate some examples and report the average time spent on each. As compensation, annotators receive $40$ pence for each annotated example.

\begin{figure}[htb!]
    \centering
    \begin{tabularx}{1\columnwidth} {
  | >{\raggedright\arraybackslash}X
  | >{\centering\arraybackslash}X
  | >{\raggedleft\arraybackslash}X |}
  \hline
  \textbf{Error Analysis}\\
  \hline
  You will be presented with pairs of reading contexts and their modified versions. The task is to compare each context and its modified version, observe the changes made and classify them into one or more of the semantic edit intention categories detailed below:
  \begin{itemize}
      \item \textit{Copy Editing}: Rephrase; improve grammar, spelling, tone, or punctuation
      \item \textit{Clarification}: Specify or explain an existing fact or meaning by example or discussion without adding new information
      \item \textit{Elaboration}: Extend/add new content; insert a fact or new meaningful assertion
      \item \textit{Fact Update}: Update numbers, dates, scores, episodes, status, etc. based on newly available information
      \item \textit{Refactoring}: Restructure the article; move and rewrite content, without changing the meaning of it
      \item \textit{Simplification}: Reduce the complexity or breadth of discussion; may remove information
      \item \textit{Vandalism}: Deliberately attempt to damage the article
      \item \textit{Other}: None of the above
  \end{itemize}
  We will use your annotation to calculate the percentage of each edit category.\\
  \hline
  \hline
  \textbf{Adversarial Validity Checking}\\
  \hline
  Please read each provided context carefully and answer a corresponding question. Select the shortest continuous span from the context as your answer. If you believe a question cannot be answered, leave the answer blank. Your answer will be compared with the ground truth answers, and the result will only be used to decide the human answerability of the question.\\
  \hline
  \end{tabularx}
    \caption{Instructions for the two distinct human annotation tasks. In the error analysis task, the eight semantic edit intentions are adopted from \citep{yang-etal-2017-identifying-semantic}.}
    \label{fig:hai}
\end{figure}

\section{Demonstration of Perturbed MRC Examples for Encoder-only Models}
\label{sec:Demonstration of Perturbed MRC Examples for Encoder-only Models}

Figure \ref{fig:dpmrceem} illustrates a naturally perturbed MRC instance each for categories C2W, C2C, and W2C, with the annotated perturbation type(s).

\definecolor{ao(english)}{rgb}{0.0, 0.5, 0.0}

\begin{figure*}[htb!]
    \centering
    \begin{tabularx}{1\textwidth} {
  | >{\raggedright\arraybackslash}X
  | >{\centering\arraybackslash}X
  | >{\raggedleft\arraybackslash}X |}
  \hline
  \textbf{Category: C2W}\\
  \hline
  \textbf{Original Paragraph:} \textit{Jacksonville, like most large cities in the United States, suffered from negative effects of rapid urban sprawl after World War II. The construction of highways led residents to move to newer housing in the suburbs. After World War II, the government of the city of Jacksonville began to increase spending to fund new public building projects in the boom that occurred after the war.} [\dots]\\
  \textbf{Perturbed Paragraph:} \textit{Jacksonville, like most large cities in the United States, suffered from negative effects of rapid urban sprawl after World War \underline{V}. The construction of highways led residents to move to newer housing in the suburbs. After World War II, the government of the city of Jacksonville began to increase spending to fund new public building projects in the boom that occurred after the war.} [\dots]\\
  \textbf{Question:} What did Jacksonville suffer from following World War I?\\
  \textbf{Prediction of \texttt{distilbert-base} and \texttt{spanbert-large-cased}:} \textcolor{ao(english)}{unanswerable}\textrightarrow\textcolor{red}{rapid urban sprawl}\\
  \textbf{Annotated Natural Perturbation Type:} Vandalism\\
  \hline
  \hline
  \textbf{Category: C2C}\\
  \hline
  \textbf{Original Paragraph:} \textit{Construction projects can suffer from preventable financial problems. Underbids happen when builders ask for too little money to complete the project. Cash flow problems exist when the present amount of funding cannot cover the current costs for labour and materials, and because they are a matter of having sufficient funds at a specific time, can arise even when the overall total is enough. Fraud is a problem in many fields, but is notoriously prevalent in the construction field. Financial planning for the project is intended to ensure that a solid plan with adequate safeguards and contingency plans are in place before the project is started and is required to ensure that the plan is properly executed over the life of the project.}\\
  \textbf{Perturbed Paragraph:} \textit{Financial planning ensures adequate safeguards and contingency plans are in place before the project is started, and ensures that the plan is properly executed over the life of the project. Construction projects can suffer from preventable financial problems. Underbids happen when builders ask for too little money to complete the project. Cash flow problems exist when the present amount of funding cannot cover the current costs for labour and materials; such problems may arise even when the overall budget is adequate, presenting a temporary issue. Fraud is also an occasional construction issue.}\\
  \textbf{Question:} What can construction projects suffer from?\\
  \textbf{Prediction of all encoder-only models:} \textcolor{ao(english)}{preventable financial problems}\textrightarrow\textcolor{ao(english)}{preventable financial problems}\\
  \textbf{Annotated Natural Perturbation Type:} Copy Editing; Refactoring; Simplification\\
  \hline
  \hline
  \textbf{Category: W2C}\\
  \hline
  \textbf{Original Paragraph:} \textit{[\dots] The antigens expressed by tumors have several sources; some are derived from oncogenic viruses like human papillomavirus, which causes cervical cancer, while others are the organism's own proteins that occur at low levels in normal cells but reach high levels in tumor cells. [\dots] A third possible source of tumor antigens are proteins normally important for regulating cell growth and survival, that commonly mutate into cancer inducing molecules called oncogenes.}\\
  \textbf{Perturbed Paragraph:} \textit{[\dots] The antigens expressed by tumors have several sources; some are derived from oncogenic viruses like human papillomavirus, which causes cancer of the cervix, vulva, vagina, penis, anus, mouth, and throat,while others are the organism's own proteins that occur at low levels in normal cells but reach high levels in tumor cells. [\dots] A third possible source of tumor antigens are proteins normally important for regulating cell growth and survival, that commonly mutate into cancer inducing molecules called oncogenes.}\\
  \textbf{Question:} What is a fourth possible source for tumor antigens?\\
  \textbf{Prediction of \texttt{bert-base-uncased}:} \textcolor{red}{proteins normally important for regulating cell growth and survival}\textrightarrow\textcolor{ao(english)}{unanswerable}\\
  \textbf{Annotated Natural Perturbation Type:} Elaboration\\
  \hline
  \end{tabularx}
    \caption{Natural perturbed MRC example in C2W, C2C and W2C categories.}
    \label{fig:dpmrceem}
\end{figure*}

\section{Process of Adversarial Validity Verification}
\label{sec:Process of Adversarial Validity Verification}

We first present two human annotators with the same collection of adversarial instances, which includes only perturbed contexts and their corresponding questions, and then ask them to answer the question based on the perturbed context. The annotators are required to select the shortest continuous span in the perturbed context that answers the question and are allowed to leave the answer blank if they are confident that the question is not answerable. Full instructions given to the annotators can be seen in Appendix \ref{sec:Human Annotation Instructions}. Subsequently, for both annotators, we measure the correctness ($1$ or $0$) of their provided answers by comparing each of them with the corresponding ground truth answers\footnote{Here, as long as one of the ground truth answers is included in the human-provided answer span, we consider the prediction to be correct.}. The inter-annotator agreement is then measured by computing the Cohen's $\kappa$ coefficient \citep{doi:10.1177/001316446002000104}. We then involve a third human annotator to annotate the adversarial examples on which the first two annotators disagree and then take the majority label as ground truth.

\section{Exhaustive Search Algorithm for Challenging Test Set Construction}
\label{sec:Exhaustive Search Algorithm for Challenging Test Set Construction}


We propose an exhaustive search algorithm that leverages the predictions of all encoder-only models to create the challenging natural perturbed test set. In detailed terms, for each matched reading passage from the prior version and its counterpart from the current version, we determine which should be designated as the \textit{original} and which as the \textit{perturbed} based on which scenario can yield the questions on which the maximum sum of the number of encoder-only models demonstrates the lack of robustness phenomenon\footnote{We define A model as lacking robustness to the perturbation if it achieves $1$ EM on the original question but attains less than $0.4$ F1 on the perturbed one (for answerable questions).}. To be specific:


Given a matched reading passage ( P ) from the prior version, its counterpart ( P' ) from the current version, and the associated questions:

\textbf{First Scenario}: We treat ( P ) as the original passage and ( P' ) as the perturbed one. We then evaluate, for each associated question, how many encoder-only models demonstrate the lack of robustness phenomenon, i.e., succeed on ( P ) but fail on ( P' ). We finally obtain the total number of models that demonstrate the lack of robustness phenomenon across all questions, denoted as ( N ). Questions on which none of the models demonstrate the lack of robustness phenomenon are removed, leaving ( Q ) questions.

\textbf{Second Scenario}: We treat ( P' ) as the original passage and ( P ) as the perturbed one. We then repeat the same evaluation process as described in the first scenario and obtain the total number of models demonstrating the lack of robustness phenomenon across all questions, denoted as ( N' ). Questions on which none of the models demonstrate the lack of robustness phenomenon are removed as well, leaving ( Q' ) questions.

If ( $N > N'$ ), we consider ( P ) as the original passage and ( P' ) as the perturbed version.

If ( $N < N'$ ), we consider ( P' ) as the original and ( P ) as the perturbed.

If ( $N = N'$ ), we compare ( Q ) and ( Q' ):

\begin{itemize}
    \item If ( $Q > Q'$ ), we consider ( P ) as the original passage and ( P' ) as the perturbed version.
    \item If ( $Q < Q'$ ), we consider ( P' ) as the original and ( P ) as the perturbed.
    \item If ( $Q = Q'$ ), the order does not matter, and we randomly decide which one should be the original and which should be the perturbed.
\end{itemize}


We finally process the identified original and perturbed passage pairs to ensure that the original passages are within the original \textsc{SQuAD 1.1} development set. For those original passages with multiple occurrences, we select the one with the maximum number of questions reserved.

\section{Natural Adversarial Samples for LLMs}
\label{sec:Natural Adversarial Samples for LLMs}

We demonstrate two naturally perturbed reading comprehension examples that pose challenges for LLMs in Figure \ref{fig:nasllms}.

\begin{figure*}[htb!]
    \centering
    \begin{tabularx}{1\textwidth} {
  | >{\raggedright\arraybackslash}X
  | >{\centering\arraybackslash}X
  | >{\raggedleft\arraybackslash}X |}
  \hline
  \textbf{\textsc{Nat\_V1\_Challenge}}\\
  \hline
  \textbf{Original Paragraph:} \textit{In business, notable alumni include Microsoft CEO Satya Nadella, Oracle Corporation founder and the third richest man in America Larry Ellison, Goldman Sachs and MF Global CEO as well as former Governor of New Jersey Jon Corzine, McKinsey \& Company founder and author of the first management accounting textbook James O. McKinsey, Arley D. Cathey, Bloomberg L.P. CEO Daniel Doctoroff, Credit Suisse CEO Brady Dougan, Morningstar, Inc. founder and CEO Joe Mansueto, Chicago Cubs owner and chairman Thomas S. Ricketts, and NBA commissioner Adam Silver.}\\
  \textbf{Perturbed Paragraph:} \textit{In business, notable alumni include Microsoft CEO Satya Nadella, Oracle Corporation founder and the third richest man in America Larry Ellison, Goldman Sachs and MF Global CEO as well as former Governor of New Jersey Jon Corzine, McKinsey \& Company founder and author of the first management accounting textbook James O. McKinsey, co-founder of the Blackstone Group Peter G. Peterson, co-founder of AQR Capital Management Cliff Asness, founder of Dimensional Fund Advisors David Booth, founder of The Carlyle Group David Rubenstein, Lazard CEO Ken Jacobs, entrepreneur David O. Sacks, CEO of TPG Group and former COO of Goldman Sachs Jon Winkelreid, former COO of Goldman Sachs Andrew Alper, billionaire investor and founder of Oaktree Capital Management Howard Marks,  Bloomberg L.P. CEO Daniel Doctoroff, Credit Suisse CEO Brady Dougan, Morningstar, Inc. founder and CEO Joe Mansueto, Chicago Cubs owner and chairman Thomas S. Ricketts, and NBA commissioner Adam Silver.}\\
  \textbf{Question:} What Goldman Sachs CEO is also an alumni of the University of Chicago?\\
  \textbf{Prediction of \texttt{GPT-3.5-turbo-0125} and \texttt{Llama-3-8B-Instruct}:} \textcolor{ao(english)}{Jon Corzine}\textrightarrow\textcolor{red}{Jon Winkelreid}\\
  \textbf{Prediction of \texttt{Falcon-40B-Instruct}:} \textcolor{ao(english)}{Jon Corzine}\textrightarrow\textcolor{red}{David Rubenstein, co-founder of The Carlyle Group, is also an alumnus of the University of Chicago.}\\
  \hline
  \hline
  \textbf{\textsc{Nat\_V2\_Challenge}}\\
  \hline
  \textbf{Original Paragraph:} \textit{Each chapter has a number of authors who are responsible for writing and editing the material. A chapter typically has two "coordinating lead authors", ten to fifteen "lead authors", and a somewhat larger number of "contributing authors". The coordinating lead authors are responsible for assembling the contributions of the other authors, ensuring that they meet stylistic and formatting requirements, and reporting to the Working Group chairs. Lead authors are responsible for writing sections of chapters. Contributing authors prepare text, graphs or data for inclusion by the lead authors.}\\
  \textbf{Perturbed Paragraph:} \textit{Each chapter has a number of authors to write and edit the material. A typical chapter has two coordinating lead authors, ten to fifteen lead authors and a larger number of contributing authors. The coordinating lead authors assemble the contributions of the other authors. They ensure that contributions meet stylistic and formatting requirements. They report to the Working Group co-chairs. Lead authors write sections of chapters. They invite contributing authors to prepare text, graphs or data for inclusion.}\\
  \textbf{Question:} Who has the responsibility for publishing materials?\\
  \textbf{Prediction of \texttt{Mistral-7B-Instruct-v0.2}:} \textcolor{ao(english)}{Unanswerable. The text does not mention any responsibility related to publishing materials.}\textrightarrow\textcolor{red}{The coordinating lead authors are responsible for publishing materials in the given context.}\\
  \hline
  \end{tabularx}
    \caption{Natural perturbed MRC examples that confuse LLMs.}
    \label{fig:nasllms}
\end{figure*}

\section{Synthetic Perturbation Methods}
\label{sec:Synthetic Perturbation Methods}

Table \ref{tab:synthetic_perturbation_techniques} presents the synthetic perturbation methods used in this study.

\begin{table*}[htb!]
    \centering
    \footnotesize
    \begin{tabular}{|p{2.7cm}p{9.3cm}|}
    \hline
    \textbf{Method}&\textbf{Description}\\
    \hline
    \multicolumn{2}{|c|}{\emph{character-level}}\\
    \hline
    CharOCR&Replace characters with Optical Character Recognition (OCR) errors.\\
    CharInsert&Inject new characters randomly.\\
    CharSubstitute&Substitute original characters randomly.\\
    CharSwapMid&Swap adjacent characters within words randomly, excluding the first and last character.\\
    CharSwapRand&Swap characters randomly without constraint.\\
    \hline\hline
    \multicolumn{2}{|c|}{\emph{word-level}}\\
    \hline
    WInsert (CWE)&Insert new words to random position according to contextual word embeddings calculation from \texttt{RoBERTa-base} \citep{DBLP:journals/corr/abs-1907-11692}.\\
    WSubstitute (CWE)&Substitute words according to contextual word embeddings calculation from \texttt{RoBERTa-base} \citep{DBLP:journals/corr/abs-1907-11692}.\\
    WSplit&Split words to two tokens randomly.\\
    WSwap&Swap adjacent words randomly.\\
    WDelete&Delete words randomly.\\
    WCrop&Remove a set of continuous word randomly.\\
    Word Synonym Substitution (WSynSub)&Substitute words with synonyms from large size English PPDB \citep{pavlick-etal-2015-ppdb}.\\
    WInsert (WE)&Insert new words to random position according to GloVe \citep{pennington-etal-2014-glove} word embeddings calculation (we use \textit{glove.6B.300d.txt}).\\
    \hline
    \end{tabular}
    \caption{Various synthetic perturbation approaches.}
    \label{tab:synthetic_perturbation_techniques}
\end{table*}

We employ methods including WSplit, WSynSub and WInsert (WE) to each sentence in the original reading passage, and then recombine the modified sentences to generate the perturbed version. Conversely, other perturbation approaches are directly executed on the entire paragraph, as implementing them at the sentence-level might result in perturbed text that is even difficult for humans to read and comprehend \citep{si-etal-2021-benchmarking}. The implementation of all character-level and word-level methods is carried out using the NLPAug library \citep{ma2019nlpaug}. Moreover, we set the perturbation rate to $30$\%, in line with the default settings within the NLPAug library.

\section{Impact of Synthetic Adversarial Training}
\label{sec:Impact of Synthetic Adversarial Training}

\begin{figure*}[htb!]
    \centering
    \includegraphics[width=\textwidth]{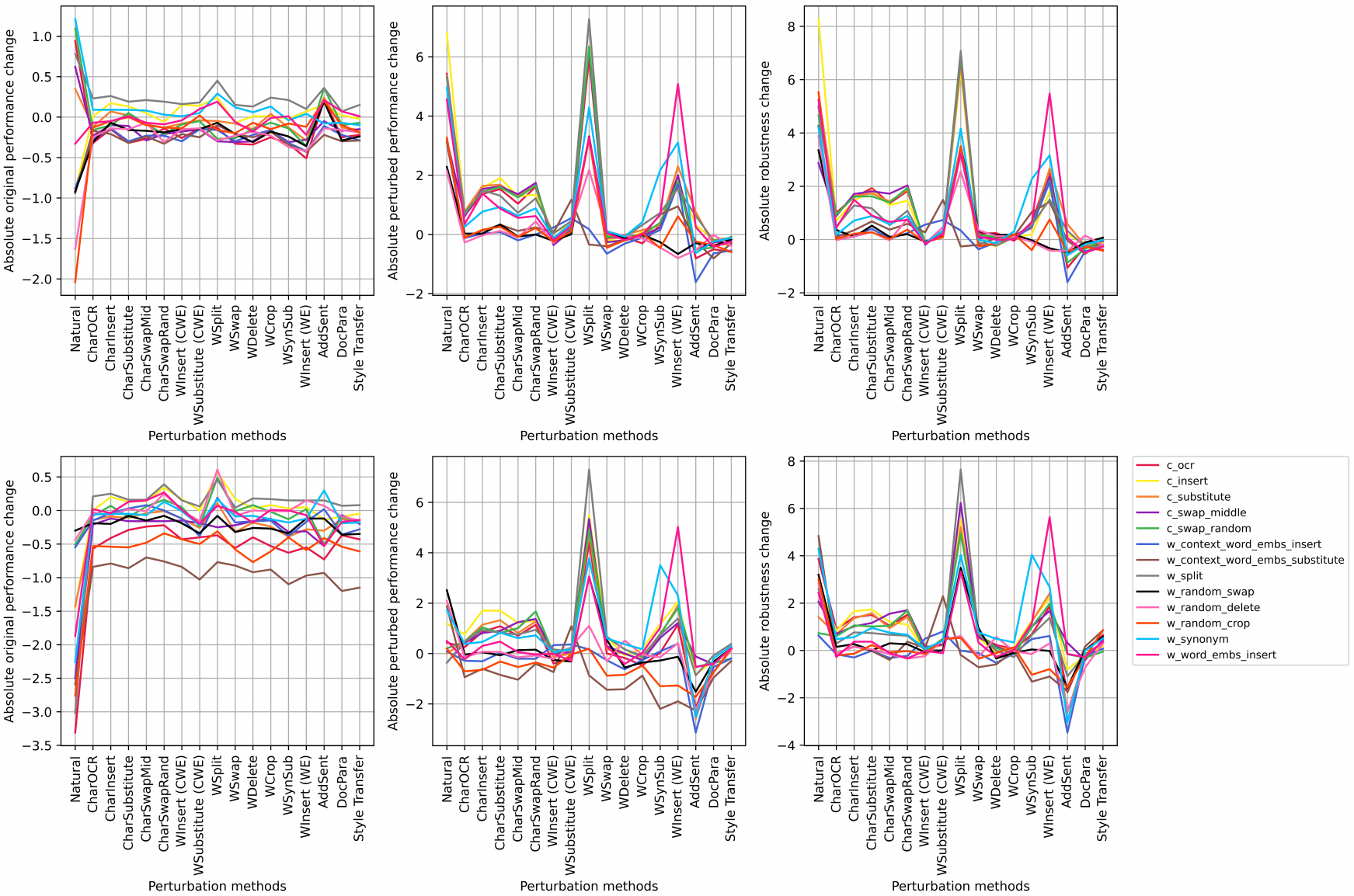}
    \caption{Absolute changes in original and perturbed performance (F1), as well as the robustness of \texttt{deberta-large} under natural and various synthetic noises, following retraining with each synthetic perturbation. The upper row and the bottom row illustrate the results on the \textsc{SQuAD 1.1} and \textsc{SQuAD 2.0} format test sets, respectively.}
    \label{fig:syn_helps_nat}
\end{figure*}

Figure \ref{fig:syn_helps_nat} describes the impact of synthetic adversarial training (for \texttt{deberta-large}) on handling natural and synthetic perturbations.

\end{document}